\documentclass{article}
\usepackage[pdftex]{graphicx}
\usepackage{amssymb}
\usepackage{spconf,amsmath,epsfig}
\usepackage{epstopdf}
\usepackage[vlined, ruled, boxed]{algorithm2e}
\SetKwInput{kwPreprocessing}{Preprocessing}
\usepackage{subfigure}
\usepackage{xspace}
\usepackage{intmacros}
\usepackage[font=small,skip=0pt]{caption}
\usepackage[belowskip=-12pt,aboveskip=0pt]{caption}
\usepackage{amsfonts}

\title{Hyperspectral Unmixing Based on Clustered Multitask Networks}
\name{Sara Khoshsokhan, Roozbeh Rajabi, Hadi Zayyani}
\address{Faculty of Electrical and Computer Engineering, Qom University of Technology, Qom, Iran}
\begin{document}
%
\maketitle
\begin{abstract}
	Hyperspectral remote sensing is a prominent research topic in data processing. Most of the spectral unmixing	algorithms are developed by adopting the linear mixing models. Nonnegative matrix factorization (NMF) and its developments are used widely for estimation of signatures and fractional abundances in the SU problem. Sparsity constraints was added to NMF, and was regularized by $ L_ {q} $ norm. In this paper, at first hyperspectral images are clustered by fuzzy c- means method, and then a new algorithm based on sparsity constrained distributed optimization is used for spectral unmixing. In the proposed algorithm, a network including clusters is employed. Each pixel in the hyperspectral images considered as a node in this network. The proposed algorithm is optimized with diffusion LMS strategy, and then the update equations for fractional abundance and signature matrices are obtained. Simulation results based on defined performance metrics illustrate advantage of the proposed algorithm in spectral unmixing of hyperspectral data compared with other methods.
\end{abstract}
\section{Introduction}
\label{sec:intro}

Remote sensing is the science of obtaining information about objects or areas from a distance, by detecting the energy that is reflected from the Earth. One of the noteworthy remote sensing techniques is hyperspectral imagery that encounter with mixed pixels challenge. Materials that are present in the scene is named endmembers. Each endmember in a pixel is weighted by its fractional abundance. The spectral unmixing (SU) technique is used to decompose a reflectance spectrum into a set of given endmember spectra. In the linear mixing model (LMM) of spectral unmixing, in which supposed that the recorded reflectance of a particular pixel, is linearly mixed by endmembers which exist in that pixel. If the number of endmembers that are present in the scene and its signatures, are unknown, the SU problem becomes a blind source separation (BSS) problem \cite{Qian11}.

Nonnegative matrix factorization (NMF) \cite{Lee01}, is a practical method of spectral unmixing, which decomposes the data into two nonnegative matrices. Recently, this basic method was developed by adding constraints, sparsity is one of the constraints for improving performance of NMF algorithm that is applied to the NMF cost function using $L_q$ regulaizers \cite{Qian11}. Regularization methods have been used to provide updating equations for signatures and abundances. Using $ L_ {1/2} $ regularization into NMF, which leads to an algorithm named  $ L_ {1/2} $-NMF, has been proposed in \cite{Qian11}, that enforces the sparsity of endmember abundances. Another approch is total variation regularized reweighted sparse NMF (TV-RSNMF) \cite{he17tvl}, that the total variation regularizer is embedded into the reweighted sparse NMF.

Recent researches have shown that the spatial autocorrelation of pixels gives beneficial information for the spectral unmixing, so in this paper, clustering methods and the distributed strategy has been used for utilization of spatial information. To do this, a clustering algorithm is applied to group all pixels into clusters across the whole hyperspectral image \cite{Ch14}, before spectral unmixing, as is done in \cite{xu17}. Then for using neighborhood information, a diffusion strategy such as least mean squares (LMS) is used because it has high stability over adaptive networks \cite{Chen14}.

To solve a distributed problem, a general network which has been named clustered multitask network \cite{Chen14}, is considered. The sparsity constrained distributed unmixing without clustering has been proposed in \cite{khosh17}. Here, we first cluster pixels using fuzzy c- means method (FCM), and then solve unmixing problem as a clustered multitask network using information of 8 neighboring pixels and sparsity constraint.

This paper is organized as follows. In section 2, we introduce the proposed method and optimize it. Section 3 provides simulation results and the last section gives conclusions.

\section{Hyperspectral Unmixing Based on Clustered Multitask Networks}
In this section, a new method that utilizes clustering of pixels and neighborhood information is proposed. First, we will express linear mixing model in subsection 2.1, and then distributed algorithm is solved for SU problem in 2.2.

\subsection{Linear Mixing Model}
To solve the SU problem, we focus on the linear mixing model (LMM). In this model, there exists a linear relation between the endmembers that weighted by their fractional abundances, in the scene. Mathematically, this model is described as:
\begin{equation}
\mathbf{y}_k=\mathbf{A}\mathbf{s}_k+\mathbf{v}_k
\end{equation}
where $\mathbf{y}_k$ is an $L\times1$ observed data vector, $\mathbf{A}$ is the $L\times c$ signature matrix, $\mathbf{s}_k=[s_k(1),s_k(2),..,s_k(c)]^T$ is the $c\times1$ fractional abundance vector and $\mathbf{v}_k$ is assumed as a $L\times1$ additive noise vector of $k$-th pixel of the image, when $c$, $L$ and $N$ denote the number of endmembers, bands and pixels, respectively.

In the SU problem, fractional abundance vectors have two constraints in each pixel, abundance sum to one constraint (ASC) and abundance nonnegativity constraint (ANC) \cite{Ma14}, which are as follows, for $c$ endmembers in a scene. 
\begin{equation}
\sum\limits_{n=1}^c \mathbf{s}_{k}(n)=1
\end{equation}
\begin{equation}
\mathbf{s}_{k}(n)\geq0,n=1,...,c
\end{equation}
Where $\mathbf{s}_{k} (n)$ is the fractional abundance of the $n$-th endmember in the $k$-th pixel of the image. Note that, in a BSS problem, only the observed vector is known and determination of two other matrices is our purpose.
\subsection{Distributed Cost Functions and Optimization}
Primarily, the fuzzy c-means clustering is adopted on the dataset \cite{bezdek84}. Then, as explained in \cite{Chen14}, three types of networks containing single task, multitask and clustered multitask networks are supposed. First, $N$ nodes are considered in a clustered multitask network and a optimum vector at node $k$ is estimated. A global cost function using LMS, $J^{global}(\mathbf{s}_k(n))$, defined as follows:
\begin{equation}
\label{eq: 4}
J^{global} (\mathbf{s}_1,\mathbf{s}_2,...,\mathbf{s}_N)=\sum\limits_{k=1}^N \mathbb{E}\{|\mathbf{y}_k -\mathbf{A} \mathbf{s}_k|^{2}\}
\end{equation}
where $\mathbb{E}$ is the expectation operator. Note that, the solution determined from global cost function, need to have access to information over all nodes, but the nodes can be considered to have availability only to information of its neighbors and the nodes of in the same cluster. Thus, the local cost function is used to solve this problem.

Then, the neighborhood information is used to turn the cost function to a distributed problem. In a distributed network, relationships between neighboring nodes are used to improve accuracy. In this article, we utilize the squared Euclidean distance \cite{Chen14}, and the $L_{q}$ regularizer for sparsity constraint that is used \cite{Qian11}. So, the following local cost function is defined, using LMS and adding the neighborhood and sparsity constraints:
\begin{equation}
\begin{aligned}
\label{eq: 10}
&J^{local} (\mathbf{s}_k)= \mathbb{E}\{|\mathbf{y}_k-\mathbf{A} \mathbf{s}_k|^{2}\}
+\eta \sum\limits_{l\in \mathcal{N}_k \cap C(k)} \rho_{kl}  ||\mathbf{s}_{k}-\mathbf{s}_{l}||^2&\\
& + \lambda ||\mathbf{s}_k||_{q}&
\end{aligned}
\end{equation}
where the $\mathcal{N}_k$ shows nodes that are in the neighborhood of node $k$, that is in the cluster $C(k)$. $\eta>0$  denotes a regularization parameter \cite{Chen14}, that controls the effect of neighborhood term, $\lambda$ is a scalar value that weights the sparsity function \cite{Qian11}, and the nonnegative coefficients $\rho_{kl}$ are normalized spectral similarity which is obtained from correlation of data vectors \cite{Chen14}:
\begin{equation}
\label{eq: lambda}
\lambda = \frac{1}{\sqrt{L}} \sum\limits_{i}  \frac{\sqrt{N}-||\mathbf{y}_i||_1/||\mathbf{y}_i||_2}{\sqrt{N-1}}
\end{equation}
\begin{equation}
\label{eq: rho}
\rho_{kj} = \frac{\theta (\mathbf{y}_k,\mathbf{y}_j)}{\sum\limits_{l \in\mathcal{N}_k^-} \theta (\mathbf{y}_k,\mathbf{y}_l)}
\end{equation}
where $\mathcal{N}_k^-$ include neighbors of node $k$ except itself, and $\theta$ is computed as \cite{Chen14}:
\begin{equation}
\label{eq: theta}
\theta (\mathbf{y}_k,\mathbf{y}_j) = \frac{\mathbf{y}_k^T \mathbf{y}_j}{||\mathbf{y}_k|| ||\mathbf{y}_j||}
\end{equation}

Now, minimizing the cost function of (\ref{eq: 10}), using the iterative steepest-descent solution \cite{Cattivelii10}, results to:
\begin{equation}
\begin{aligned}
\label{eq: s}
&\mathbf{s}_k^{i+1} = P^+\big[ \mathbf{s}_k^{i}+ \mu \mathbf{A}^T (\mathbf{y}_k-\mathbf{A}\mathbf{s}_k)&\\
&- \mu \eta \sum\limits_{l\in \mathcal{N}_k\cap C(k)} \rho_{kl} (\mathbf{s}_l^{i}-\mathbf{s}_k^{i})&\\
&- \mu \lambda \frac{\big( \mathbf{s}_k^{i}\big) |\mathbf{s}_k^{i}|^{q-2}}{||\mathbf{s}_k^{i}||_{q}^{q-1}} \big] &
\end{aligned}
\end{equation}
where $\mu>0$ is a step-size parameter, $i$ denotes the iteration number, and $P^+$ operator projects vectors onto a simplex, that adopt the ASC and ANC constraints for abundance vectors. This operator explained in \cite{Chen11}.

Also, the spectral signatures matrix is updated similar to NMF algorithm, using multiplicative update rules \cite{Lee01}:
\begin{equation}
\label{eq: A}
\mathbf{A}^{i+1}=\mathbf{A}^i*\frac{\mathbf{Y}\mathbf{S}^T} {\mathbf{A}\mathbf{S}\mathbf{S}^T} 
\end{equation}

A significant point in implementation of the algorithm is stopping criteria. This approach will be stopped until the maximum number of iteration ($T$), or the following stopping criteria is reached.
\begin{equation}
\label{eq: stop}
||J_{new}-J_{old}||<\epsilon
\end{equation}
where $J_{new}$ and $J_{old}$ are cost function values for two consecutive iterations and $\epsilon$ has been set to $10^{-8}$ in our experiments. The proposed approach is summarized in Algorithm 1.
\begin{algorithm}
	\SetKwInOut{Input}{input}
	\SetKwInOut{Output}{output}
	\SetKwInput{Initialization}{Initialisation}
	\Input{Hyperspectral data matrix ($\mathbf{Y}$)\\
		Parameters: $C$,$N$,$c$,$L$,$q$,$\mu$ and $\eta$,}
	\Output{Estimated fractional abundance and signature matrices ($\mathbf{S}$ and $\mathbf{A}$),}
	\kwPreprocessing{Clustering $\mathbf{Y}$ using FCM algorithm to $C$ clusters, determines $C(k)$, $k=1,...,N$,}
	\Initialization{Initialise the $\mathbf{A}$ and $\mathbf{S}$ matrices by random matrices or the outcome of VCA algorithm \cite{Nascimento05}. Compute  $\rho$ values from (\ref{eq: rho}),}
	\While{the maximum number of iteration ($T$) or stopping criteria in (\ref{eq: stop}) has been reached,}
	{
		a. Update $\mathbf{A}$, using  (\ref{eq: A})\;
		b. Update $\mathbf{s}_k$ for all pixels, by applying  (\ref{eq: s})\;
		c. Adopt $P^+$ operator for ASC and ANC constraints\;
		end}
	\caption{Hyperspectral Unmixing Based on Clustered Multitask Networks}
	\label{algorithm}
\end{algorithm}

\section{Experiments and Results}
In this section, for demonstration of quantitative comparison between the proposed method with different number of clusters, also between the proposed and other methods, the performance metrics such as spectral angle distance ($SAD$) and abundance angle distance ($AAD$) are used.

\begin{figure}[tb]
	\centering
	\includegraphics[width=1.9in]{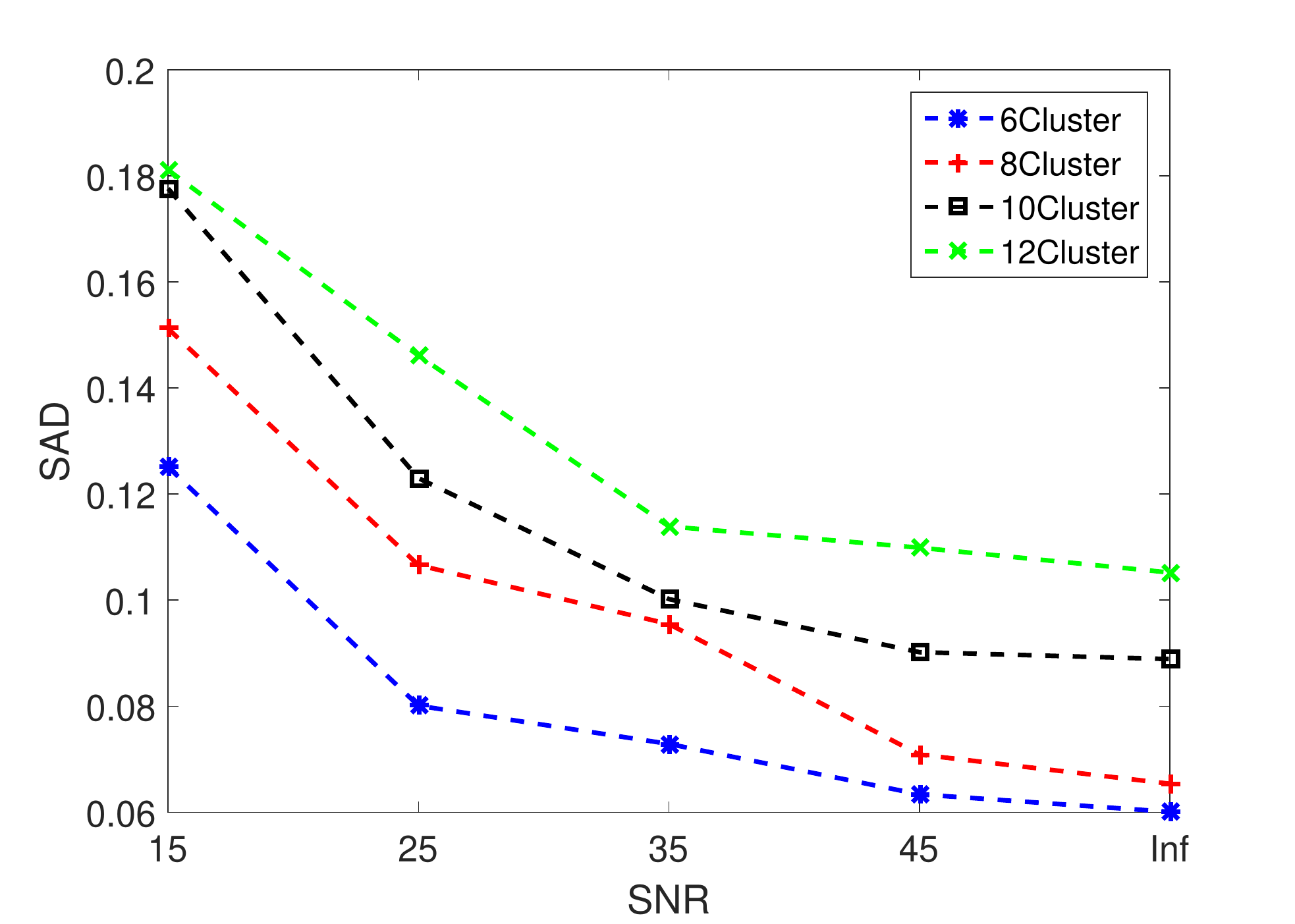}
	\caption{The SAD performance metric of the proposed algorithm applied on synthetic dataset for 6 endmembers, with different number of clusters and using VCA initialization.}
	\label{cl}
\end{figure}

\begin{figure}[tb]
	\centering
	\subfigure[]{
		\includegraphics[width=4cm]{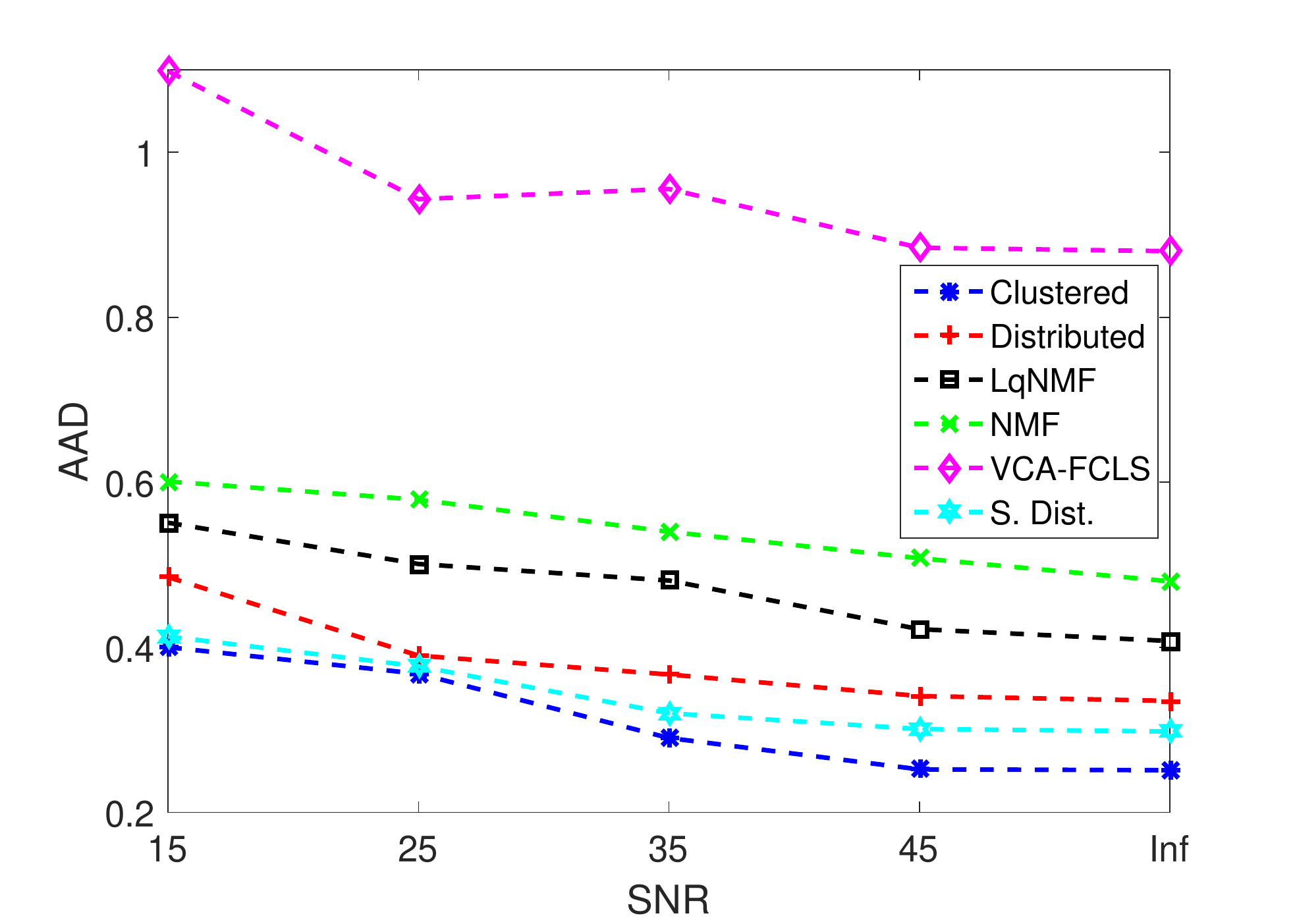}
		\label{fig:s1}
	}
	\subfigure[]{
		\includegraphics[width=4cm]{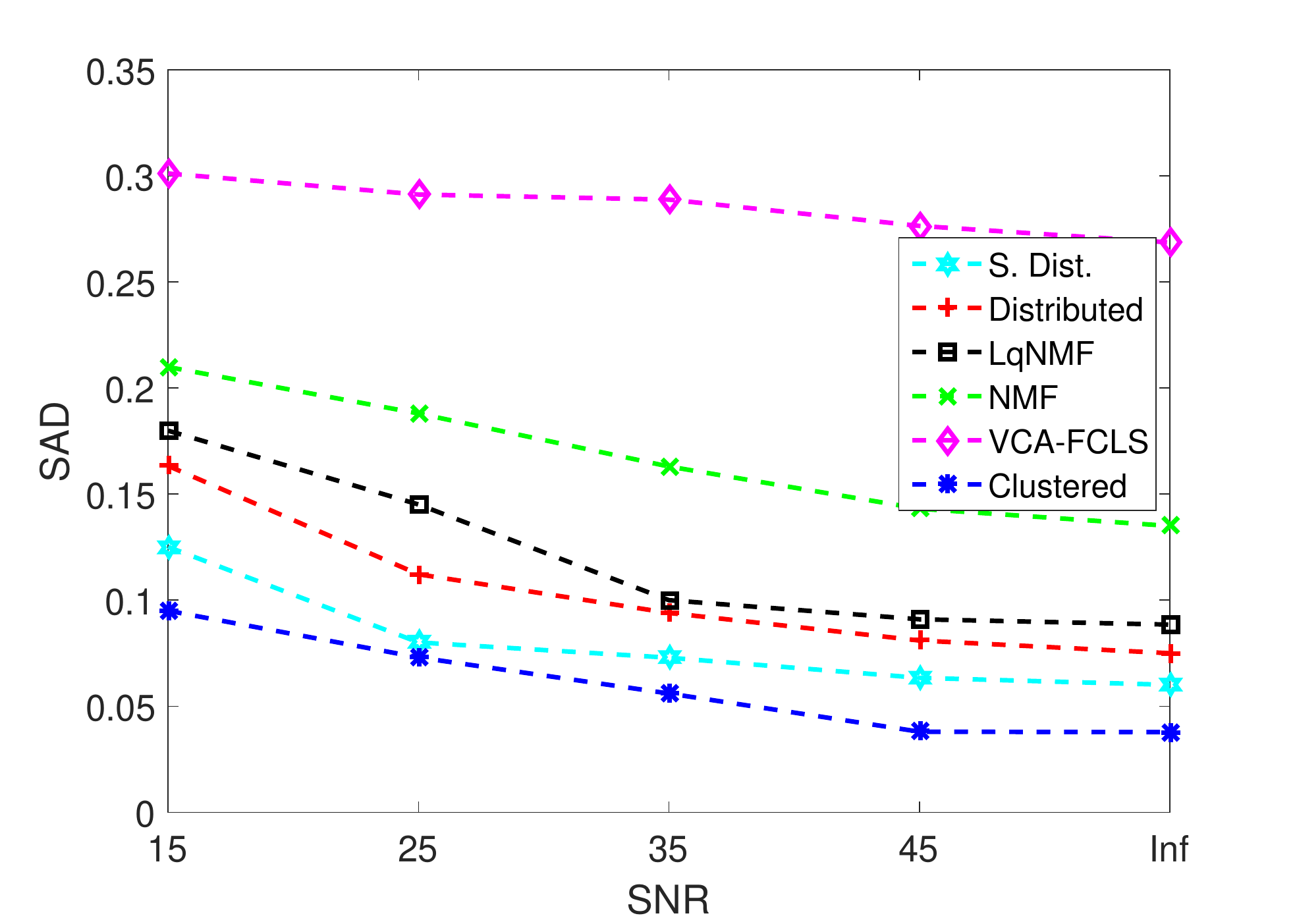}
		\label{fig:2}
	}
	\caption{(a)The $AAD$ and (b) $SAD$ performance metric of 6 methods versus SNR, using VCA initialization and applied on synthetic data. }
	\label{methods}
\end{figure}

\begin{figure}[tb]
	\begin{center}$
		\begin{array}{lll}
		\includegraphics[width=20mm]{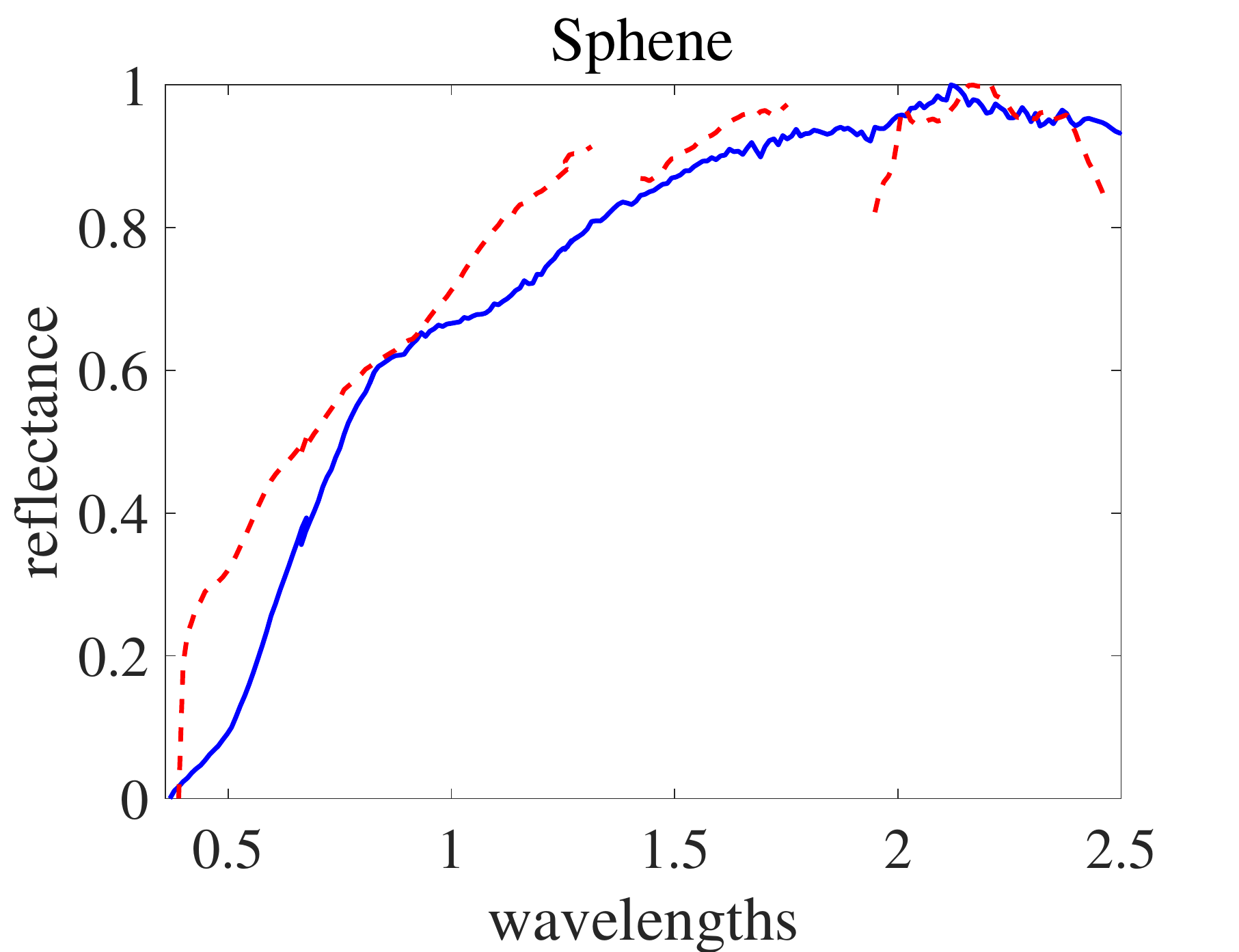}&
		\includegraphics[width=20mm]{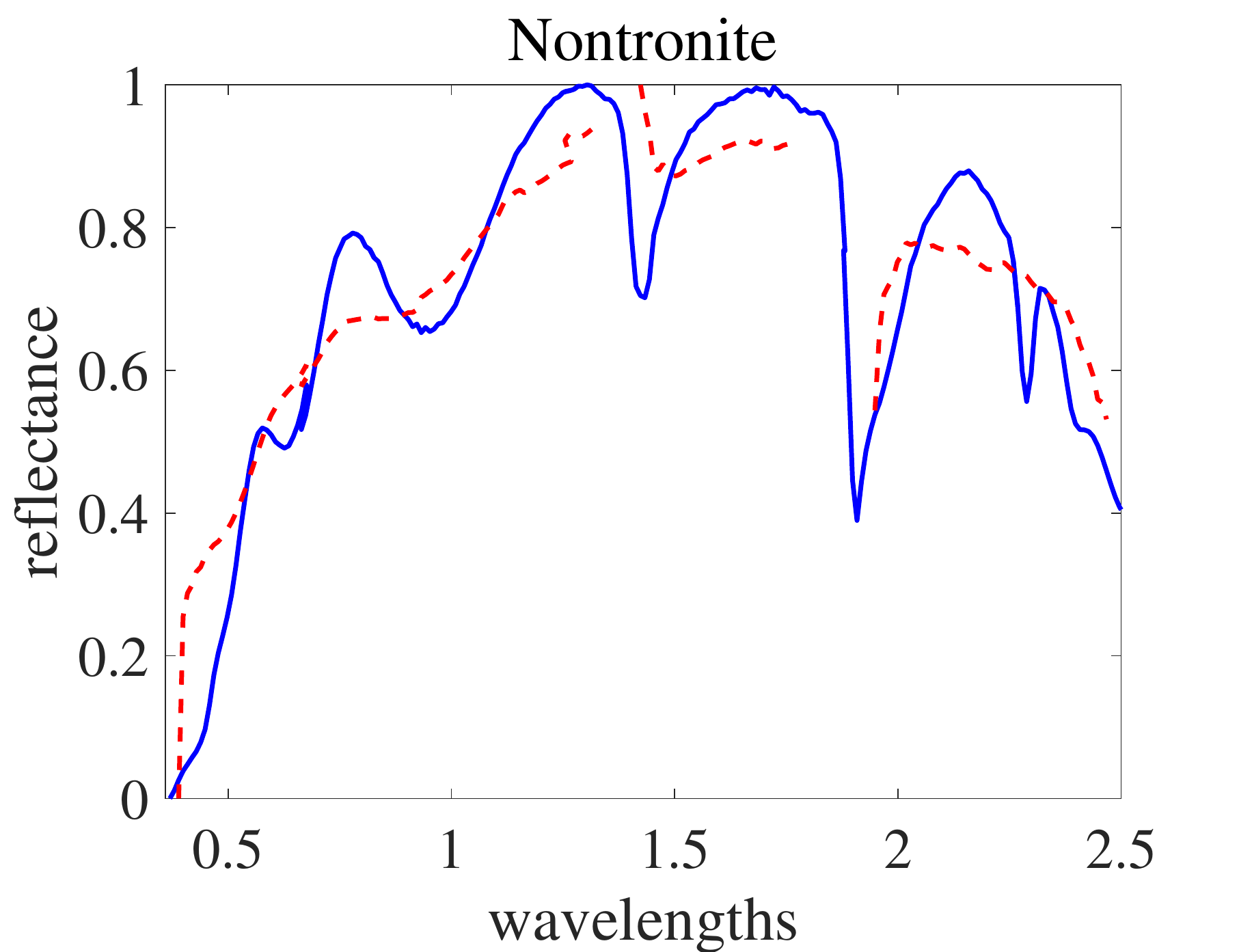}&
		\includegraphics[width=20mm]{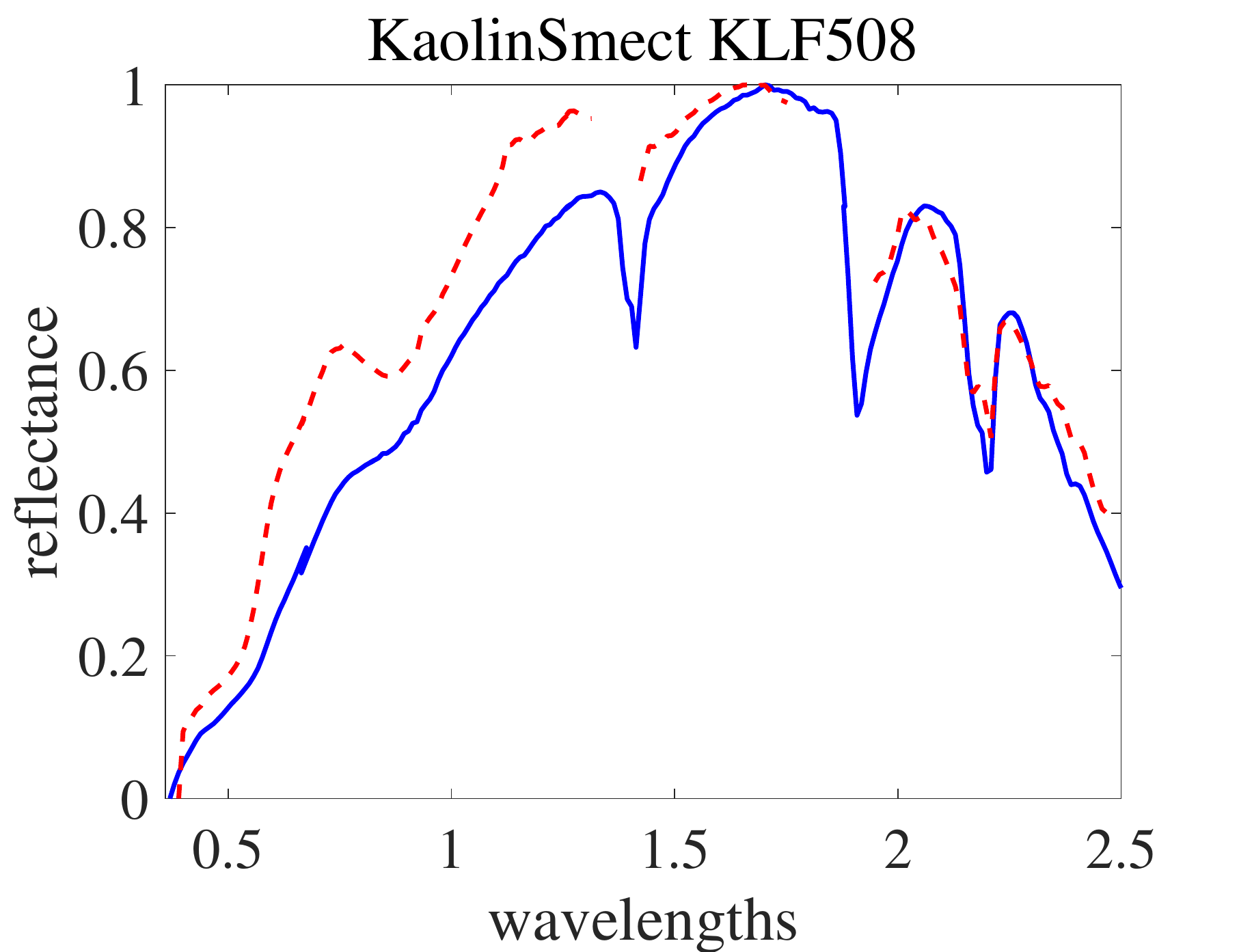}
		\end{array}$
	\end{center}
	
	\begin{center}$
		\begin{array}{lll}
		\includegraphics[width=20mm]{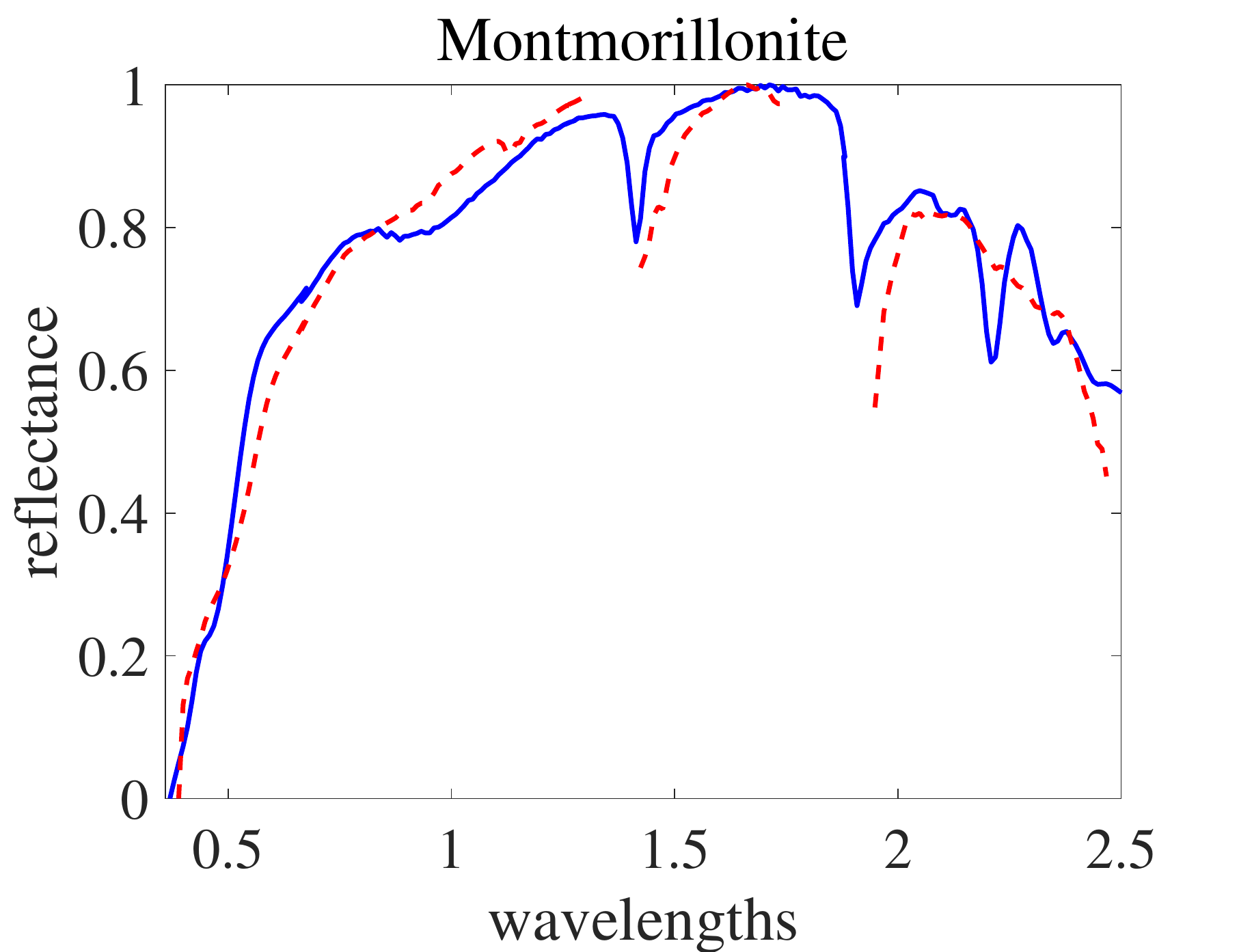}&
		\includegraphics[width=20mm]{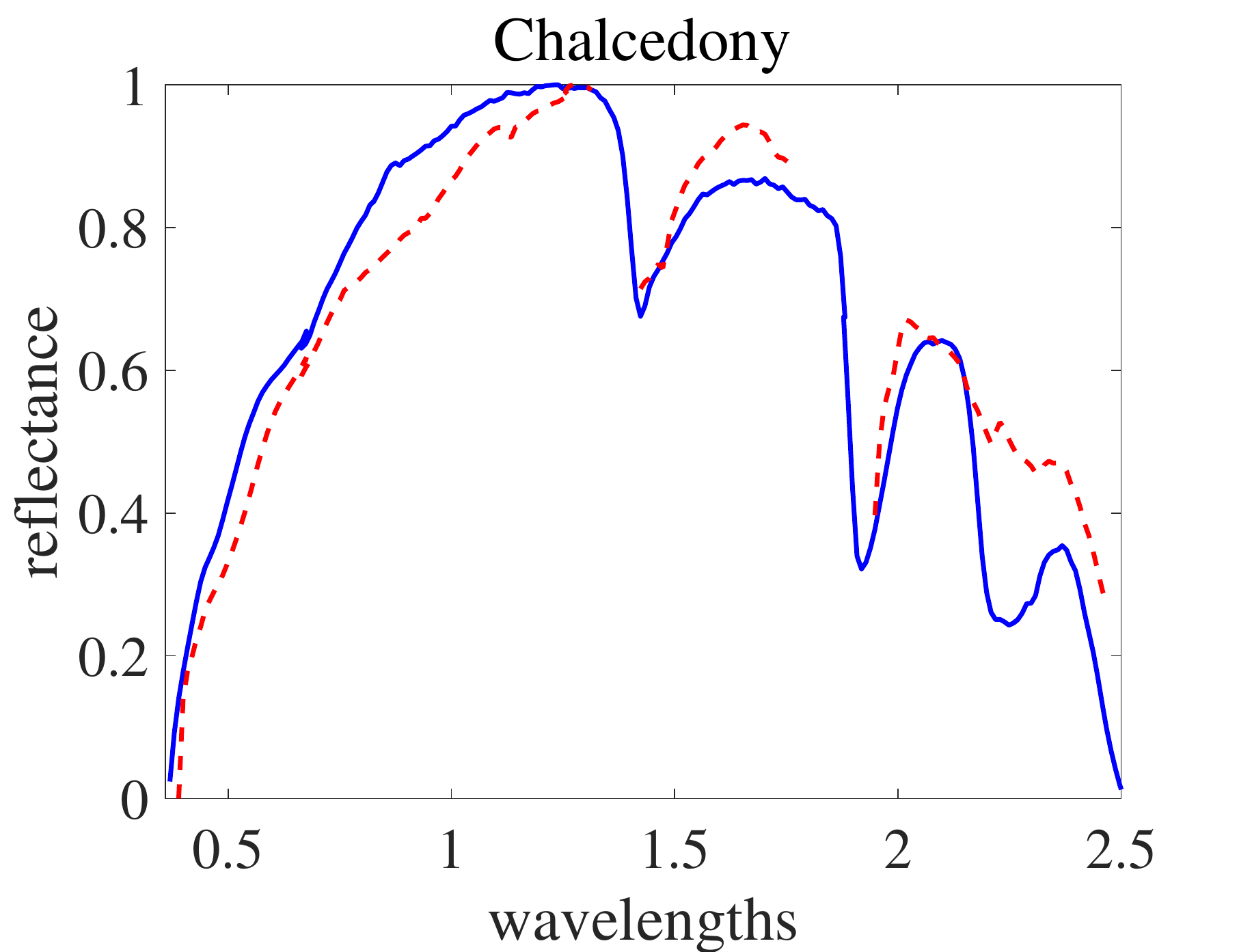}&
		\includegraphics[width=20mm]{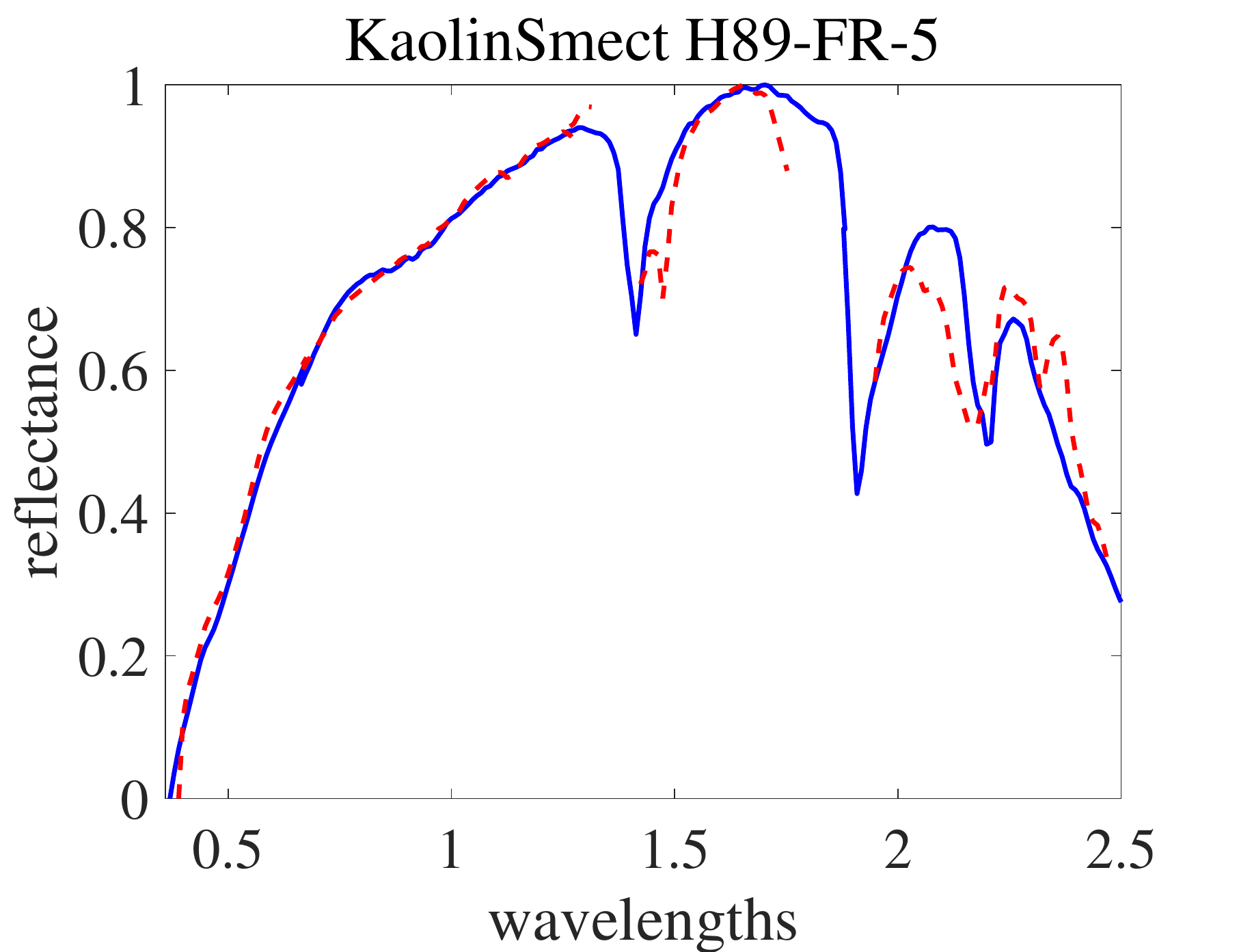}
		\end{array}$
	\end{center}
	
	\begin{center}$
		\begin{array}{lll}
		\includegraphics[width=20mm]{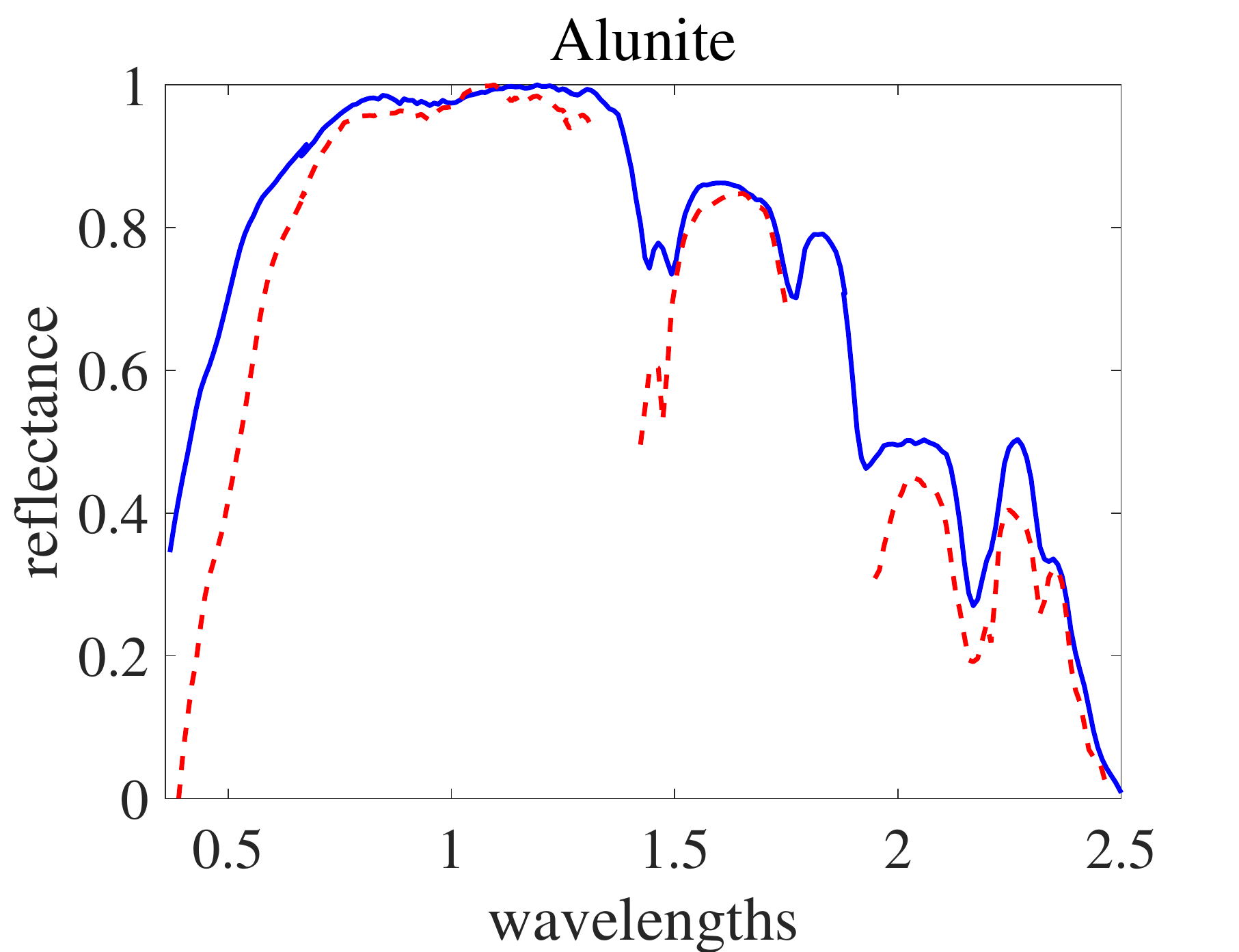}&
		\includegraphics[width=20mm]{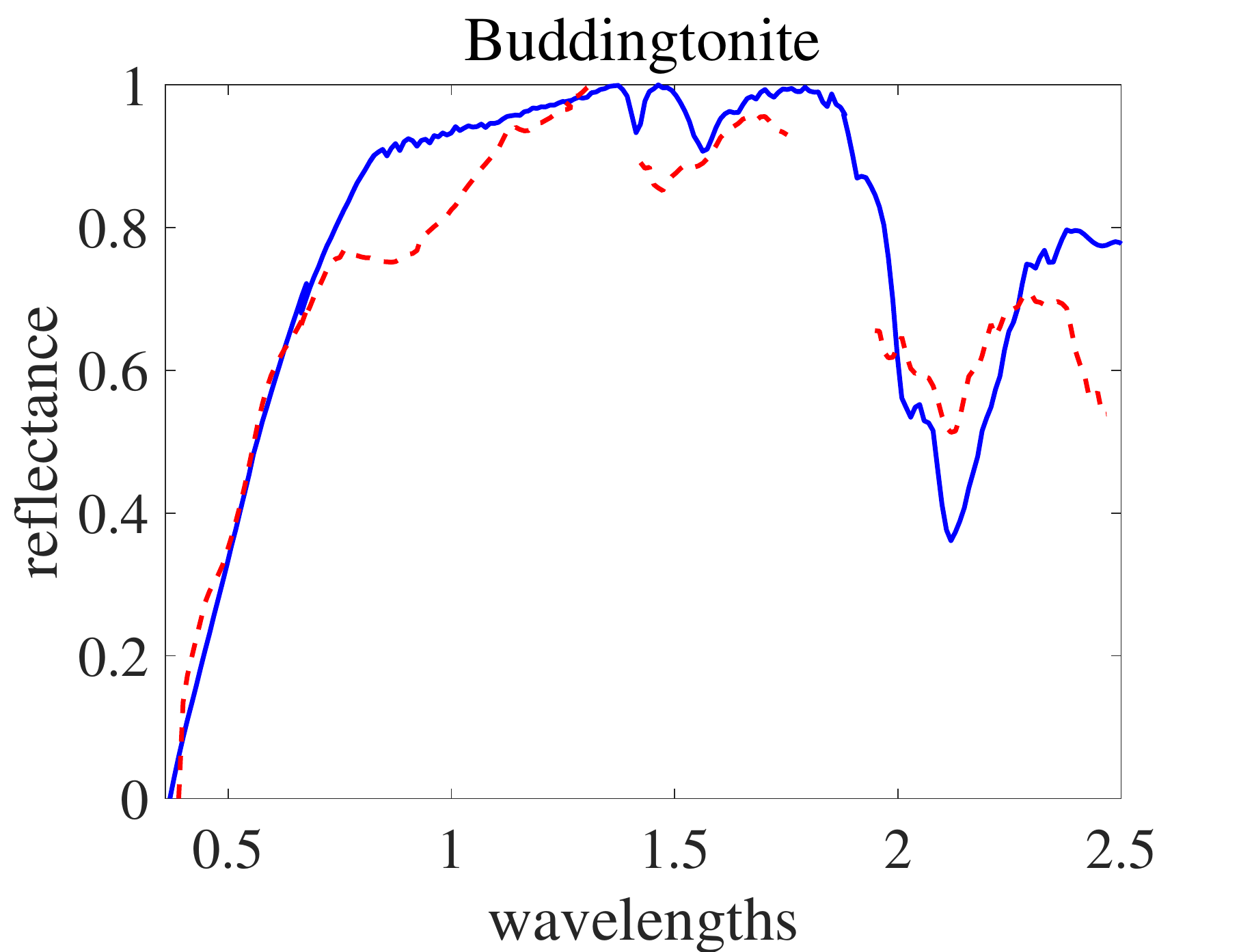}&
		\includegraphics[width=20mm]{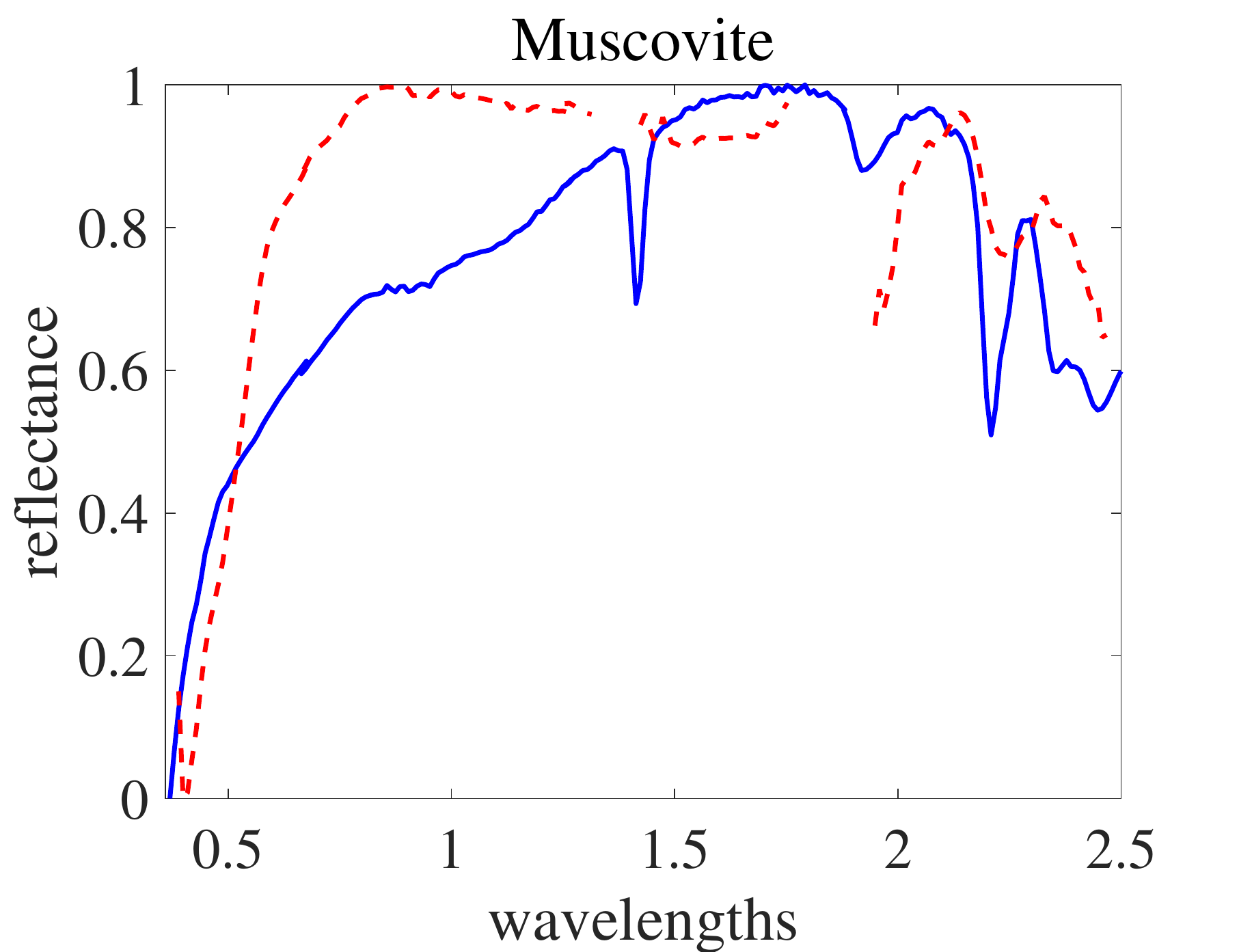}
		\end{array}$
	\end{center}
	
	\begin{center}$
		\begin{array}{lll}
		\includegraphics[width=20mm]{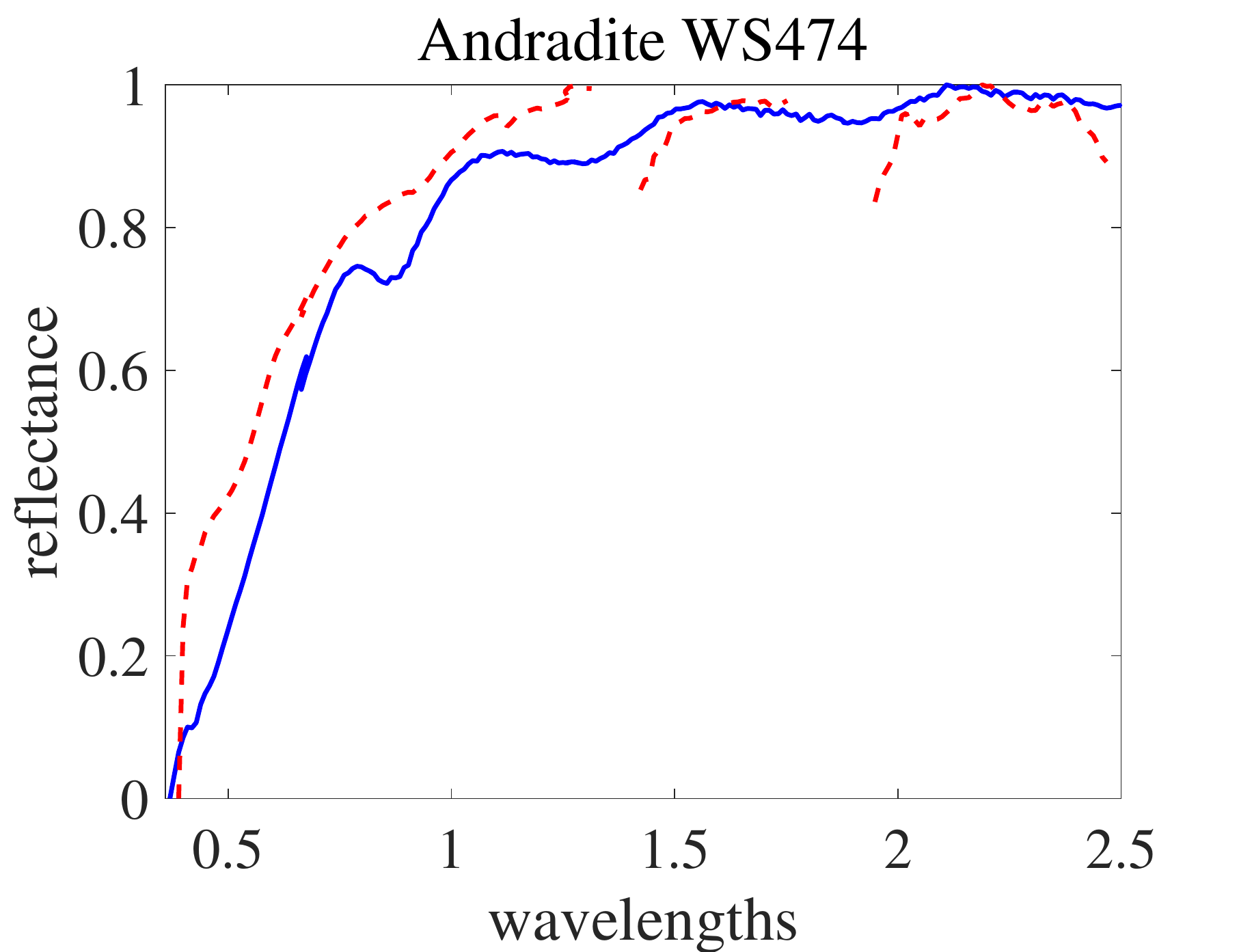}&
		\includegraphics[width=20mm]{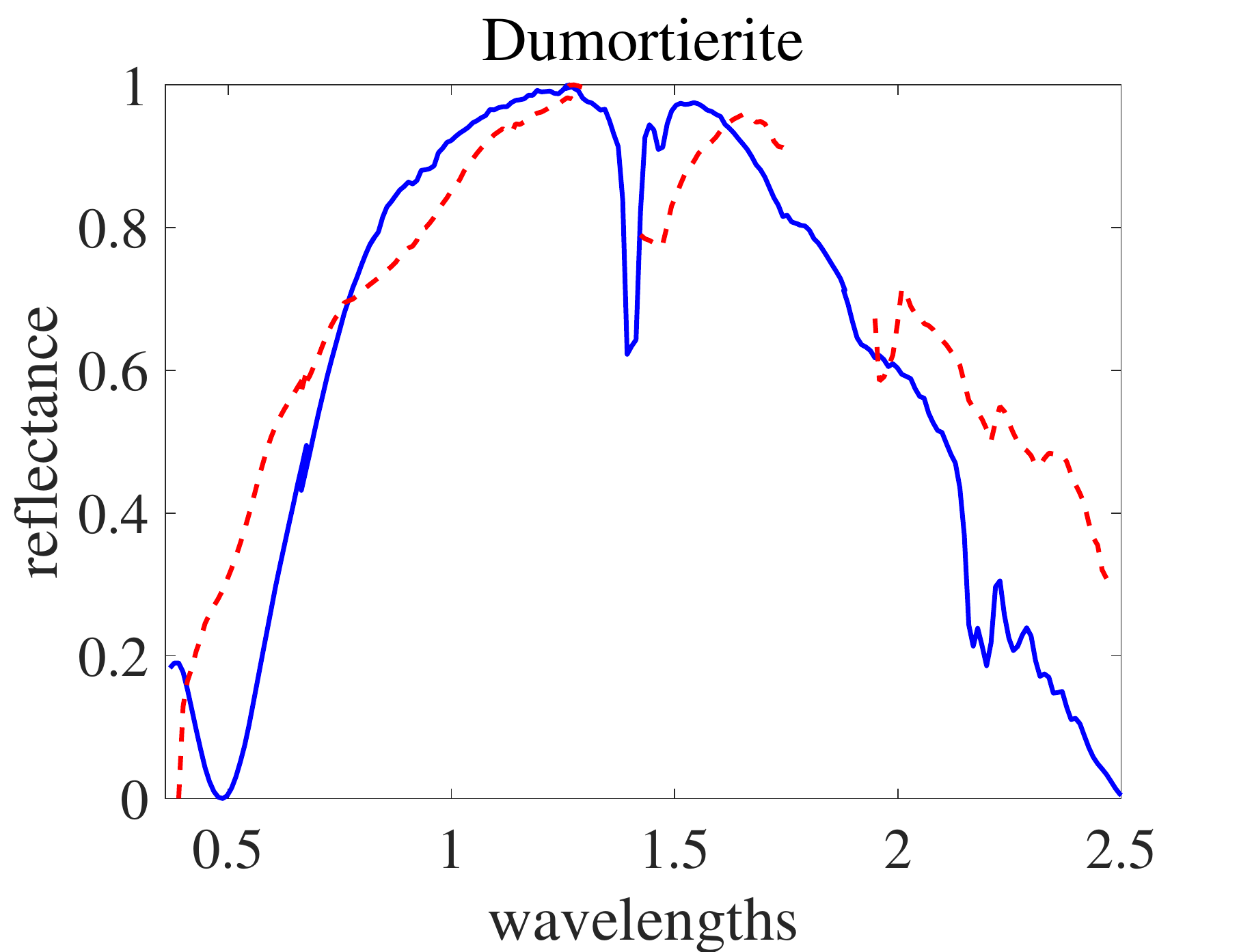}&
		\includegraphics[width=20mm]{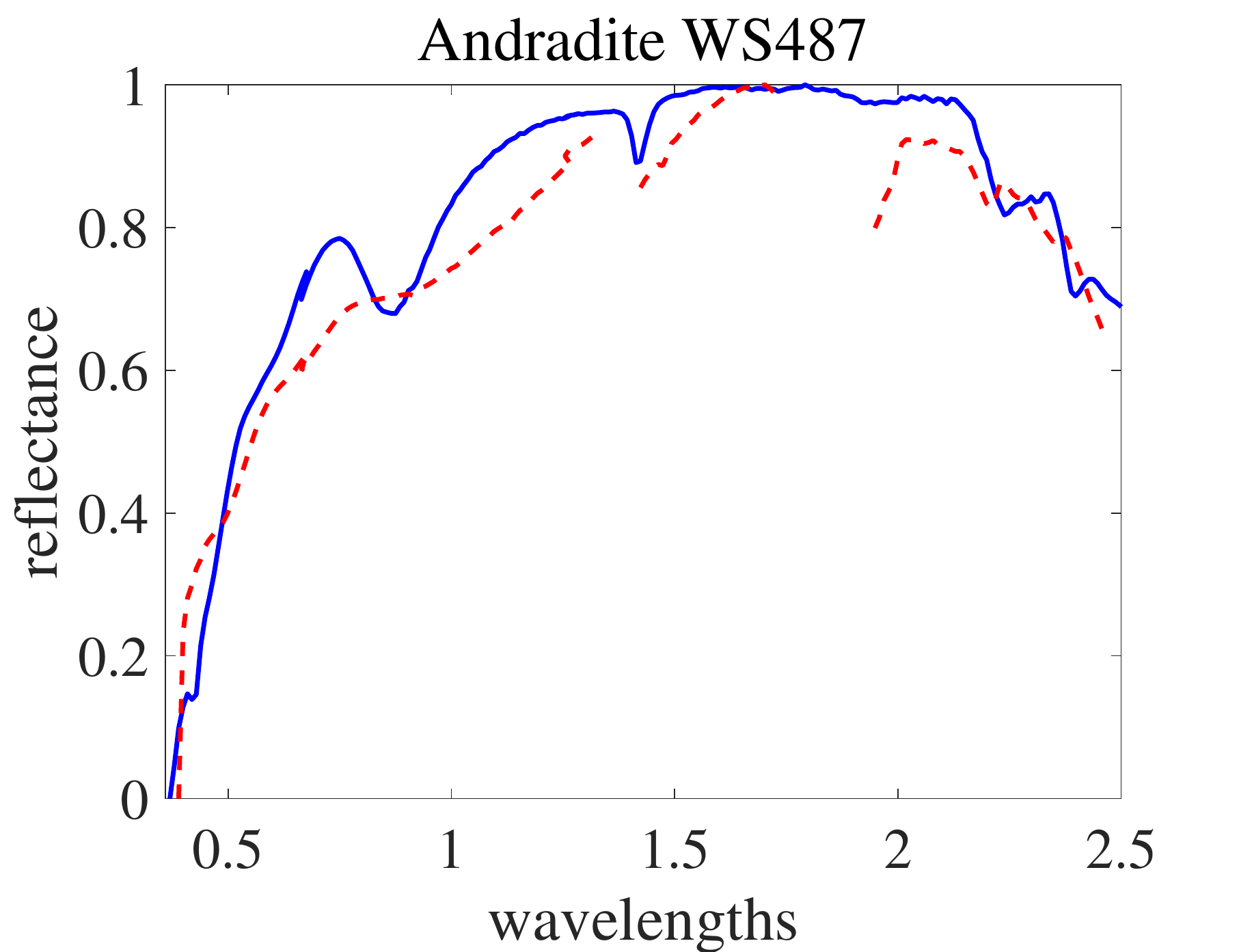}
		\end{array}$
	\end{center}
	\caption{Original spectral signatures (blue solid lines) and estimated signatures of proposed algorithm (red dashed lines) versus wavelengths ($\mu m$), on AVIRIS Cuprite dataset and using VCA initialization.}
	\label{real}
\end{figure}

\begin{figure}[tb]
	\begin{center}$
		\begin{array}{lll}
		\includegraphics[width=20mm]{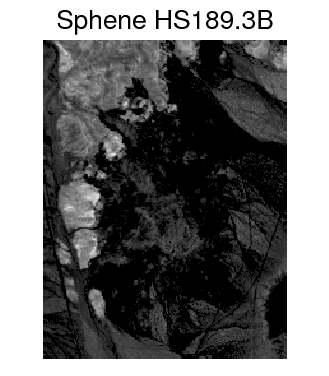}&
		\includegraphics[width=20mm]{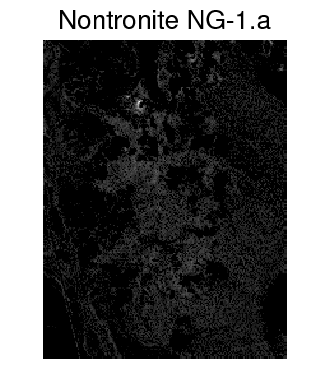}&
		\includegraphics[width=20mm]{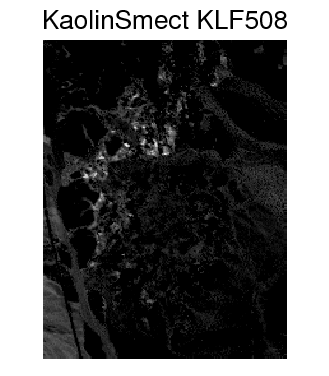}
		\end{array}$
	\end{center}
	\begin{center}$
		\begin{array}{lll}
		\includegraphics[width=20mm]{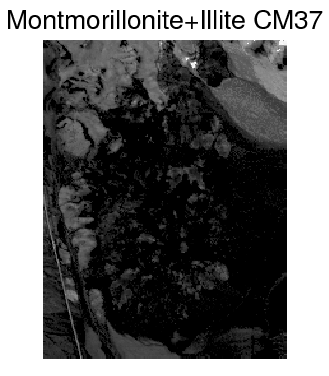}&
		\includegraphics[width=20mm]{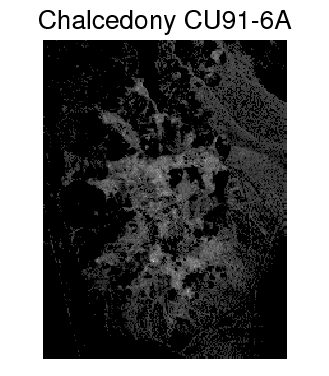}&
		\includegraphics[width=20mm]{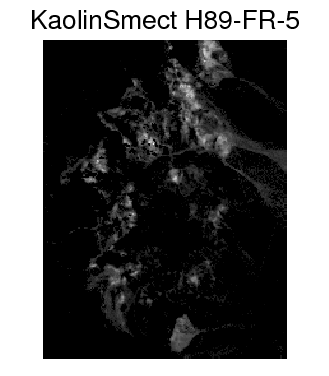}
		\end{array}$
	\end{center}
	\begin{center}$
		\begin{array}{lll}
		\includegraphics[width=20mm]{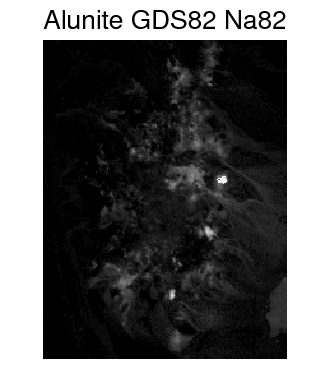}&
		\includegraphics[width=20mm]{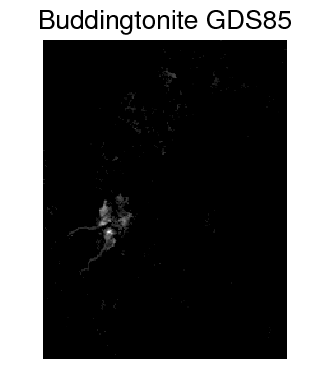}&
		\includegraphics[width=20mm]{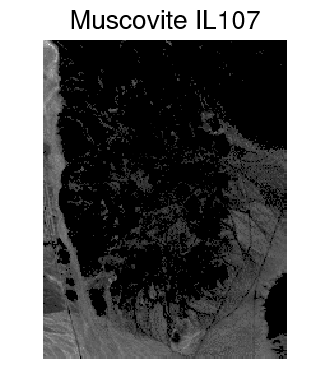}
		\end{array}$
	\end{center}
	\begin{center}$
		\begin{array}{lll}
		\includegraphics[width=20mm]{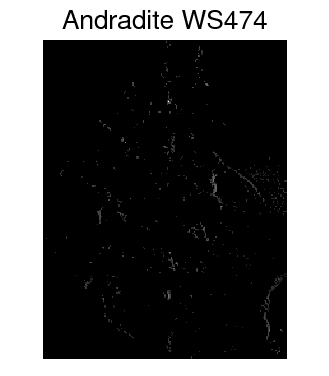}&
		\includegraphics[width=20mm]{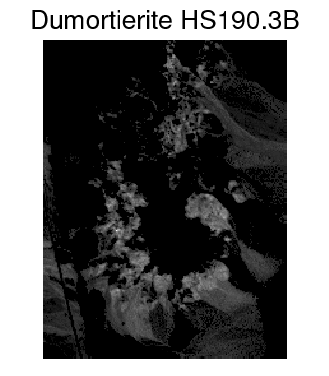}&
		\includegraphics[width=20mm]{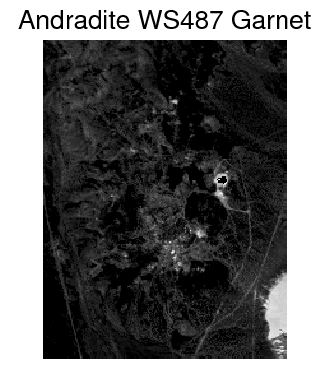}
		\end{array}$
	\end{center}
	\caption{Estimated fractional abundances of endmembers that are present in the AVIRIS Cuprite data scene, using sparsity constrained distributed unmixing and VCA initialization.}
	\label{real1}
\end{figure}

Primarily, the proposed algorithm has been applied on synthetic data. to generate this dataset, six signatures of USGS library have been selected randomly, using a 7$\times$7 low pass filter and containing no pure pixels. Then, the zero mean Gaussian noise with 5 different levels of SNR have been added to generated data, and performance metrics have been computed by averaging 20 Monte-Carlo runs. To choose the best number of clusters in our experiments, the SAD performance metric has been evaluated, and then according to \figurename{~\ref{cl}}, the best number of clusters has been set to 6, that is equal to number of endmembers. Also, values of $\mu$ and $\eta$ has been considered equal to 0.02 and 0.1, respectively \cite{Chen14}, and $q=1$, to gain the best results. Then the proposed algorithm and some other algorithms such as VCA-FCLS \cite{Nascimento05}, NMF \cite{Lee01}, $L_{1/2}$-NMF \cite{Qian11}, distributed unmixing \cite{Chen14}, sparsity constrained distributed unmixing \cite{khosh17} and TV-RSNMF \cite{he17tvl}, that is similar to the proposed algorithm without clustering step, has been applied on the generated synthetic dataset. The comparison of performance metrics of this 7 different methods has been shown in \figurename{~\ref{methods}}, where the metrics of proposed algorithm is star-dashed line and excels other methods. Afterwards, the proposed algorithm has been applied on AVIRIS Cuprite real dataset \cite{Green98}. After clustering into 12 clusters, the simulation results of spectral signatures and fractional abundances have been shown in \figurename{~\ref{real}} and {~\ref{real1}}. Also, $SAD$ performance metric of $L_{1/2}$-NMF, distributed unmixing, sparsity constrained distributed unmixing and proposed method on the real dataset have been compared in \tablename{~\ref{table1}}, the results of proposed algorithm are available in the last column and has the best $rmsSAD$ value.

\begin{table}[tb]
	\centering
	\caption{
		The $SAD$ performance metric of five algorithms on AVIRIS Cuprite dataset, using VCA initialization.
	}
	\resizebox{\columnwidth}{!}{%
		\begin{tabular}{c|c|c|c|c|c}
			\hline
			materials & $L_{1/2}$-NMF & Dist. & S. Dist. &  TV-RSNMF & Proposed \\
			\hline \hline
			Sphene &0.2143 & \textbf{0.1561} & 0.1673 &0.1583 & 0.1574\\
			\hline
			Nontronite  & 0.2518 & 0.1944 & 0.1743 & 0.1803&\textbf{0.1711}\\
			\hline
			KaolinSmect \#1 & \textbf{0.1653} & 0.2370 & 0.1741 & 0.1731 & 0.1702\\
			\hline
			Montmorillonite & 0.2318 & 0.3571 & \textbf{0.2103}& 0.2159 & 0.2248 \\
			\hline
			Chalcedony & 0.1995 & 0.1603 & 0.1653& 0.1588 & \textbf{0.1437}\\
			\hline
			KaolinSmect \#2 &\textbf{0.2542} & 0.2873 & 0.2608 & 0.2576 & 0.2596\\
			\hline
			Alunite &0.3458 & 0.3813 & \textbf{0.2369} & 0.2551 & 0.2417\\
			\hline
			Buddingtonite & 0.1693 & 0.2514 & 0.1953 & 0.2034 & \textbf{0.1643}\\
			\hline
			Muscovite & 0.1584 & 0.4682 & \textbf{0.1537} & 0.1563 & 0.1575\\
			\hline
			Andradite \#1 &0.3361 & \textbf{0.2132} & 0.2425 & 0.2392& 0.2337\\
			\hline
			Dumortierite & \textbf{0.2453} & 0.3381 & 0.2639 & 0.2686 & 0.2519\\
			\hline
			Andradite \#2  & 0.3829 & 0.3711 & 0.2854 & 0.3136 &\textbf{0.2472}\\
			\hline\hline
			rmsSAD  & 0.2562 & 0.2998 & 0.2153 & 0.2207 & \textbf{0.2064}\\
			\hline
		\end{tabular}
	}
	\label{table1}
\end{table}

\section{Conclusion}
Spectral unmixing (SU) is a technique to characterize mixed pixels in hyperspectral images measured by remote sensors. Decomposition of pixels in the scene into their constituent materials is the goal of spectral unmixing. This paper followed two steps, including the FCM clustering of hyperspectral images and then the sparsity constrained distributed unmixing method. This new algorithm considered sparsity, clustering and neighborhood information. Simulation results on synthetic and real datasets illustrated preference of proposed approach in comparison against NMF, $L_{1/2}$-NMF, VCA-FCLS, distributed unmixing and sparsity constrained distributed unmixing methods.


\bibliographystyle{ieeetr}
\bibliography{tran}

\end{document}